\documentclass{ieeeaccess}
\usepackage{cite}
\usepackage{amsmath,amssymb,amsfonts}
\usepackage{algorithmic}
\usepackage{graphicx}
\usepackage{textcomp}
\usepackage{xurl}
\usepackage{tabularx}
\usepackage{arydshln}
\usepackage{multicol, multirow, booktabs, makecell}
\usepackage{longtable, dashrule}
\usepackage{placeins}
\usepackage{microtype}
\usepackage{subcaption}
\usepackage[colorlinks=true, linkcolor=black, urlcolor=black, citecolor=black]{hyperref}
\def\BibTeX{{\rm B\kern-.05em{\sc i\kern-.025em b}\kern-.08em
    T\kern-.1667em\lower.7ex\hbox{E}\kern-.125emX}}

\usepackage[T1]{fontenc}
\usepackage[scaled]{helvet}

\begin{document}
\history{Date of publication not available yet, date of current version June 26, 2026.}
\doi{not available yet}

\title{Optimizing Image Preparation and Compression for Face Recognition within 1024 Bytes}
\author{\uppercase{Paul Andreas\authorrefmark{1},
Torsten Schlett\authorrefmark{1}, and Christoph Busch\authorrefmark{1}
\IEEEmembership{Fellow, IEEE}}}
\address[1]{Department of Computer Science, Hochschule Darmstadt, Schöfferstraße 3, 64295, Darmstadt, Germany\\ (e-mail: paul@samen-andreas.de, torsten.schlett@h-da.de, christoph.busch@h-da.de)}
\tfootnote{This research work has been funded by the German Federal Ministry of Education and Research and the Hessian Ministry of Higher Education, Research, Science and the Arts within their joint support of the National Research Center for Applied Cybersecurity ATHENE.}

\markboth
{Andreas \headeretal: Optimizing Image Preparation and Compression for Face Recognition within 1024 Bytes}
{Andreas \headeretal: Optimizing Image Preparation and Compression for Face Recognition within 1024 Bytes}
\corresp{Corresponding author: Paul Andreas (e-mail: paul@samen-andreas.de).}

\begin{abstract}
ICAO-compliant machine readable travel documents enable automated biometric face verification. The biometric reference is stored on an RFID chip included in form of a JPEG or JPEG 2000 compressed facial image. In contrast, temporary travel documents lack of machine readability, which excludes the owner from such automated processes. This disadvantage could be solved by equipping such documents with 2D barcodes. This technology offers a resource-saving alternative to expensive RFID chips, while still offering machine readability and fast issuing processes. However, this solution introduces the challenge of storing the face images at significantly smaller storage capacities, creating the need for reducing the file size of the included facial image to a maximum of 1024 bytes. This study examines preprocessing steps and compression configurations, using JPEG, JPEG 2000, JPEG XL, JPEG AI, HEIF, AVIF, and WebP for image compression to this target size, while still preserving as much face recognition performance as possible. While the reference sample must always comply with ICAO specifications, the individual samples may or may not meet these requirements, depending on the application. This work optimizes compression steps for both of these prerequisites. It is shown that the recently standardised JPEG AI, when using optimized settings, provides the best face recognition performance, in particular when the comparison includes only images with high face image quality. AVIF and WebP also provide good results in both scenarios. The losses caused by the strong lossy compression are comparatively small. For the comparison of ICAO-compliant face images only, converting the images to grayscale proves to be a helpful preprocessing step, whereas for comparisons involving less suitable samples, preserving color is preferable. In addition, smoothing and resizing the images beforehand also turns out to be beneficial.
\end{abstract}


\titlepgskip=-15pt

\maketitle

\section{Introduction}
\label{sec:introduction}
Biometrics nowadays are a common way to identify or authorize people. Most smartphones support unlocking by finger print or face recognition, ID cards contain portraits of the owner, and even cars are able to recognize the current driver to adjust settings accordingly. The advantages of biometrics are obvious: In contrast to classical keys or passwords one cannot forget or lose a biometric trait, and it cannot be shared with unauthorized third parties. A particularly convenient biometric characteristic is the face. Humans naturally are good at recognizing faces. This enables keeping the human in the loop. In addition, the face can easily be captured from a distance, which further increases user acceptance. As such, with advancing digitalization it has great potential to be included in more processes, especially in public places where identification is needed. One example for such a place would be an airport. Here, thousands of people must be identified every day to ensure safety. One step towards more digitalized processes are Automated Border Control (ABC) gates. \\
As their name suggests, they automate the control process at border crossings. In order to successfully prove travelers identity here, they need an identity document with an RFID chip that contains a biometric, standard-compliant facial image. This requirement for an ABC gate has the consequence that every user needs a Machine Readable Travel Document (MRTD) \cite{icao_doc9303_2021}. Considering temporary travel documents, these must either be equipped with an RFID chip — which is neither an economical option due to their short period of use nor practical since this massively increases the issuing time — or another option must be implemented to integrate a compliant photograph into the document, so that it is still machine readable. A common and already standardized solution for this are 2D barcodes \cite{iso_iec_18004_2024}\cite{iso_iec_16022_2024}\cite{iso_iec_24778_2024}. Here, the challenge is to encode the ICAO compliant portrait photo with sufficient quality, since 2D barcodes offer only limited storage capacities. \\
To overcome this challenge, this study examines and evaluates various settings in the context of compressing facial images. These include:
\begin{itemize}
    \item Selection of the best compression algorithm: The study examines the suitability of JPEG, JPEG 2000, JPEG XL, JPEG AI, AVIF, HEIF, and WebP.
    \item Evaluation of the optimal parameters for the selected compression algorithm: Each compression algorithm comes with individual settings for fine tuning the compression behavior.
    \item Choosing an optimal resolution: Using lower resolutions results in smaller file sizes. Reducing the resolution does delete information, but it does not add any special compression artifacts. In our study we determine an optimal ratio of resolution to compression rate so that as little information as possible is distorted.
    \item Manipulating the input image for better compression algorithm performance: Some areas of a face image are more important than others. By deleting some information in these less important areas, the overall face recognition performance might increase.
    \item Deciding between color or grayscale images: Color images can commonly be encoded with three color channels, whereby each channel describes the amount of red, green, and blue for each pixel. Grayscale images only use one channel, reducing the bits used per pixel by two thirds without any other form of compression. However, by converting a color image to a grayscale image, information is discarded, which could decrease the face recognition performance more than the use of a higher compression ratio setting later on.
\end{itemize}
For this optimization, a process is presented and implemented that extracts the feature vectors of the facial images before and after compression, to measure the impact of compression on the faces. The goal is to find those settings that have the least impact on biometric features of the face image. After the optimization, the influence of compression on the system’s face recognition performance is evaluated. The aim is to determine whether encoding facial images in 2D barcodes is feasible at all in our application scenarios, and, if so, which measures need to be taken for this purpose.

\subsection{Relationship to Prior Publication}
\label{subsec:prior_work}
This work is an extension of our preliminary study published recently \cite{IWBF2026}. The novelty of this work is the additional focus on the application scenario of ABC gates. Since ABC gates are designed so that the biometric probe sample taken on site is of the highest possible face image quality, for this scenario a comparison of the compressed images can be made only against images of high face image quality, meaning that probe images with (strong) pose are of less relevance in this extended study. Due to this constraint, the optimized parameters turn out to be fundamentally different, which in turn also influences the results and the recommendations derived from them. \\
In addition, this version provides further insights into the impact of the various operations on the similarity scores, detailed information on the parameter optimization process, and the results of this optimization for all examined compression algorithms.  

\subsection{Structure of this paper}
Section \ref{sec:related_work} discusses research work that has already carried out similar studies in the biometric field. In Section \ref{sec:exp_setup}, we then present the data and software used, as well as our implementations and the specific procedure for optimizing facial image compression. In Section \ref{sec:results}, we address the insights that result from our experiments. These include the optimization strategies identified, the impact of compression on the faces themselves, as well as the effects on the overall performance of face recognition and face image quality. Finally, Section \ref{sec:results} provides a summary of the paper.

\section{Related work}
\label{sec:related_work}
This section provides an overview of existing work that addresses research questions or objectives similar to those pursued in this study. As in this study, the effects of lossy compression on the performance of biometric systems were investigated. Up to now, the main focus has been on the different performance levels of various compression algorithms. One aspect of the investigation that distinguishes this work from existing studies is its additional focus on the preprocessing of the data to be compressed. \\
A basic prerequisite that is essential in this work is the target size of 1024 bytes for the compressed face images. A study that also followed this approach is technical report NIST SP 500-343 by Grother et al.\@ \cite{grother_face_2025}. That study considers QR Codes as storage medium. For this purpose, the authors examined target file sizes of 600, 800, 960, 1040, and 1200 bytes. The authors investigate the suitability of the compression algorithms JPEG, JPEG 2000, JPEG LI, JPEG XL, HEIC, AVIF, and WebP. They found that JPEG and JPEG LI are unsuitable for such a purpose, as they are often unable to achieve the specified target sizes with sufficiently preserved details to facilitate effective facial recognition. Subsequently, Grother et al.\@ examined the error rates of the compressed images for the different combinations of resolution, target size, and compression algorithm and found that from a target size of 960 bytes upward, error rates similar to those of uncompressed images become achievable. The compression algorithms found to be the most suitable for this were WebP, AVIF, HEIC, and JPEG 2000. Overall, the authors observed that the mated similarity scores decrease on average due to the compression of the images. Another finding of the authors is that reducing the resolution has less negative impact on face recognition performance than stronger lossy compression encoder settings, which is why a low resolution of the images is recommended. \\
Another study that deals with the face recognition performance of images compressed to predefined target file sizes is the work of Schlett et al.\@ \cite{effectlossycompressionalgorithms}. In this study, the authors examined the compression algorithms JPEG, JPEG 2000, and JPEG XL, as well as the reduction of the file size of PNG images by rescaling them to target sizes of 5 kB, 4.5 kB, 4 kB, 3.5 kB, 3 kB, 2.5 kB, and 2.2 kB. Here, the authors analyzed both the face recognition performance of the compressed images relative to one another and the evolution of face image quality. Out of the examined methods, JPEG XL turned out to be the most suitable algorithm for face image compression without severe losses in face recognition performance. Moreover, only JPEG was found to be unsuitable for particularly small target sizes, which is consistent with the findings of Grother et al.\@ \cite{grother_face_2025}. With regard to face image quality, the authors observed a slight reduction, which was comparatively minor. \\
In the recent investigation of Bousnina et al.\@ \cite{deep_learning_impact}, the authors examined the differences in the impact of conventional compression algorithms compared to AI-based compression algorithms on the face recognition performance of ArcFace. To this end, the authors compared the effects of JPEG, JPEG 2000, and JPEG XL with those of Ballé \cite{balle}, Cheng \cite{cheng}, and the HiFiC codec \cite{hific}. They evaluated the peak signal-to-noise ratios, the multi-scale structural similarity index measure, as well as False Match Rate (FMR) and False Non-Match Rate (FNMR). It turned out that the AI-based compression algorithms provided both slightly better image quality and improved face recognition performance at the same compression rate compared to the conventional algorithms. Another finding of the authors was that, although face recognition performance was generally negatively affected by compression, in some cases an improvement occurred. They attributed this to the noise-filtering properties of the compression algorithms, which, at medium to high compression rates, may have the effect that the compressed images are less noisy, which in turn would benefit face recognition performance. \\
A study that applied the approach of manipulating the images before the main compression encoding, is the work of Maser et al.\@ \cite{finger_vein_compression}, albeit in the context of finger vein images. The aim of the authors was to replace the background of the images with a uniform gray, and to investigate the resulting effects on compression for lossy and lossless algorithms, as well as on recognition accuracy in the case of lossy compression. For lossless compression, lossless JPEG, JPEG XR, GIF, PNG, JPEG LS and ZIP were used; for lossy compression, JPEG, JPEG 2000, JPEG XR and BPG were examined. \\
The results of the study showed that replacing the background with a uniform gray improved the compression performance of lossless algorithms. However, this effect could not be transferred to lossy compression. In this case, recognition performance was often negatively affected by this manipulation of the original image. The authors’ explanation for this phenomenon lies in the strong edges between background and foreground that were introduced by this operation. Along these edges, lossy compression led to the formation of artifacts, which in turn had a negative impact on recognition accuracy.

\section{Experimental Setup}
\label{sec:exp_setup}
This chapter describes the details of our procedure to optimize the compression settings and the additional prior image processing. For this purpose, two sub-datasets are prepared, all images are preprocessed for face recognition, and a collection of parameters is presented that is used for optimization. In addition, a process is designed that enables the comparison of the influence the parameters have on the compressed results and the resulting face recognition performance. The source code used for this study is available at: \\ \textbf{\url{https://github.com/dasec/1kB-FaceImage}}.

\subsection{Dataset}
\label{subsec:dataset}
For evaluation the ColorFERET database \cite{phillips1998feret}\cite{phillips2000feret} is used. This dataset consists of a total of 11,128 uncompressed facial images belonging to 966 individuals. Images of each individual can be included from different perspectives. These can include frontal views, slight rotation, side profiles, with neutral and other facial expressions, different clothing, different backgrounds, different hairstyles, as well as different eye levels. For reasons of better readability and standardization, this entire dataset will be referred to as the “full dataset” in the following pages.\\
Since images used for compression must later on meet the requirements for an identification document, it is not feasible to use all images of the dataset for compression evaluation. These identification document images must be taken under certain conditions, as stated in international standards ISO/IEC 19794-5 \cite{iso_iec_19794_5_2011} and ISO/IEC 39794-5 \cite{iso_iec_39794_5_2019} and as demanded by the International Civil Aviation Organization (ICAO) \cite{icao_doc9303_2021}. Consequently, only a subset of the original dataset containing exclusively frontal images is used to evaluate the effects of compression on face recognition. This subset will be referred to as the “frontal dataset” in the rest of this paper. It comprises 2,638 images. Not all of the used images show strictly neutral facial expressions, since the provided ColorFERET data lacked labels to easily enable such a selection. \\ 
Since the search for optimal parameters and further preprocessing steps is iterative, the frontal dataset must be reduced further. I.e.\@ because the various adjustment parameters can be combined with each other, there are too many possibilities to process all frontal face images per algorithm each time, given computational resource constraints of this study at least. For this reason, another subset is created for the search, which contains only 10 images of 10 individuals (one image per individual). The images for this dataset are manually chosen to be as diverse as possible. This subset will be referred to as the “parameter search dataset”. Based on the parameter search dataset, all effects of the settings are examined in the following. Afterwards only the best three settings are validated on the frontal dataset.

\subsection{Used Face Image Processing Software}
\label{subsec:software_used}
In order to biometrically process the facial images, we use various software. This includes the Face Image Quality Assessment Toolkit (FIQAT) \cite{fiqat}, which provides various functions from image preprocessing and feature extraction through to quality assessment. In doing so, it makes use of a number of external models and libraries. For face detection, including facial landmarks, we use a Sample and Computation Redistribution for Efficient Face Detection (SCRFD) \cite{scrfd} model from the InsightFace project \cite{insightface_ai_2026}. \\ 
For face recognition feature extraction we use an openly available state-of-the-art model from the CVLface library \cite{cvlface}, which uses a ViT-KPRPE architecture \cite{kprpe} trained with AdaFace \cite{adaface} loss on WebFace12M \cite{webface12m}. This model will be referred to as “AdaFace” herein. For feature vector comparison we correspondingly use the cosine similarity in order to estimate the facial similarity. \\
In addition to FIQAT and AdaFace, we use the Open Face Image Quality (OFIQ) software library \cite{ofiqlib}\cite{Merkle-OFIQ-Report-240930}, which is maintained by the German Federal Office for Information Security. Here, we use an algorithm for face region estimation, and in a subsequent step the OFIQ Unified Quality Score (UQS) model to assess the face image quality, the latter being based on MagFace \cite{magface}. Further, we also use a more recently developed Face Image Quality Assessment (FIQA) model based on Vision Transformer architectures, called ViT-FIQA(C) \cite{vitfiqa}.

\subsection{Image preparation}
\label{subsec:img_prep}
\begin{figure}[!htbp]
    \centering    
    \begin{subfigure}[b]{0.25\textwidth}
        \centering
        \includegraphics[width=\textwidth]{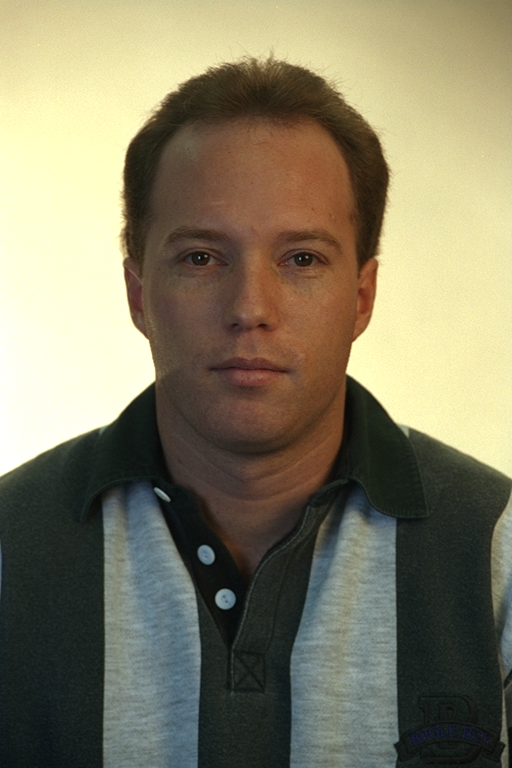}
        \caption{}
        \label{fig:original}
    \end{subfigure}
    \hspace{0.2cm}
    \begin{subfigure}[b]{0.18\textwidth}
        \centering
        \includegraphics[width=\textwidth]{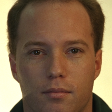}
        \caption{}
        \label{fig:cropped}
    \end{subfigure}    
    \caption{This figure shows the original image included in ColorFERET (\ref{fig:original}) and the resulting image obtained after cropping and alignment (\ref{fig:cropped}). Note: The size ratios of the two images to each other do not match the real ones, because the image from ColorFERET has a very high resolution.}
    \label{fig:setup:crop_and_align}
\end{figure}

In order to reduce the face image file size to 1024 bytes or less, different preparation steps are taken. First, the relevant face area of each image is cropped out and aligned. This is always a necessary step for contemporary face recognition algorithms and already reduces the file size by a significant amount, since the majority of unnecessary information in the background of the larger original image is deleted in this step. We are using FIQAT for this task. The specific facial region is cut out in three steps. First, the faces in an image are detected using SCRFD \cite{scrfd}. Then the face that the algorithm considers to be the most reliable is selected, but in the used ColorFERET data there always is only one face in the image regardless. In the next step, this face is aligned and cut out using the five landmarks (corners of the mouth, center of the eyes, and tip of the nose) detected by SCRFD \cite{scrfd}, whereby the resolution is adjusted to 112×112 pixels required by the AdaFace model. The effects of this image processing are shown in Fig. \ref{fig:setup:crop_and_align}. 
Since in later processing steps the compressed images have to be compared with the image that would be used in a comparison without any file size limitation (e.g.\@ a live capture), this version is stored as the original for all future image variants. Adaptations to different resolutions are always carried out by the native function of FIQAT. 

\subsection{Parameters for Optimization}
\label{subsec:params_for_optimization}
Parameters for optimization are grouped into five categories. These consist of: 
\begin{itemize}
    \item Compression algorithm
    \item Settings for compression algorithm
    \item Color
    \item Resolution
    \item Image manipulation
\end{itemize}
The following sections describe the individual parameters used for the optimization.

\subsubsection{Compression algorithms}
\label{subsubsec:compr_algorithms}
We are evaluating seven different compression algorithms. These are JPEG, JPEG 2000, JPEG XL, JPEG AI, AVIF, HEIF, and WebP. The exact versions used including the underlying compression codecs are listed in Table \ref{tab:compression_algorithms}.
\begin{table}[!htbp]
    \setlength{\tabcolsep}{4pt}
    \centering
    \renewcommand{\arraystretch}{1.2}
    \caption{Information about the compression algorithms used.}
    \begin{tabular}{lccc}
    \hline
         \thead{Compression \\ algorithm} & Software & Version & Codec \\ 
         \hline
         JPEG & libjpeg-turbo \cite{libjpeg-turbo} & 3.1.1 & \\ 
         JPEG 2000 & OpenJPEG \cite{openjpeg} & 2.5.4 & \\
         JPEG XL & libjxl \cite{libjxl} & 0.11.1 & \\
         JPEG AI & JPEG AI Reference \cite{jpegai-ref-software} & - & Commit \href{https://gitlab.com/wg1/jpeg-ai/jpeg-ai-reference-software/-/tree/c11b7c1c869aeebd031efb84402d07a76d4d4e15}{c11b7c1c}\\
         AVIF & libavif \cite{libavif} & 1.3.0 & AOM v3.13.1 \cite{libaom} \\
         HEIF & libheif \cite{libheif} & 1.21.0 & H.265 v3.5 \cite{libx265} \\
         WebP & libwebp \cite{libwebp} & 1.6.0 & \\
         \hline
    \end{tabular}
    \label{tab:compression_algorithms}
\end{table}

\subsubsection{Settings for compression algorithm}
\label{subsubsec:cli_settings}
Each compression algorithm offers individual settings for specifying the compression process. The lists of the parameters evaluated per algorithm are shown in Table \ref{tab:compression_algorithm_settings}.
\begin{table}[!htbp]
    \centering
    \caption{Included compression settings per algorithm.}
    \begin{tabular}{lcc}
    \hline
         \thead{Compression \\ algorithm} & CLI parameter & Value range used \\ 
         \hline
         & quality & 0-100 \\
         & grayscale & - \\
         & rgb & - \\
         JPEG & smooth & 0-100 \\
         & optimize & - \\
         & arithmetic & - \\
         & progressive & - \\
         \hdashline
         & ratio & 0-100 \\
         & quality & - \\
         & number of resolutions & 1-7 \\
         & block height & 32, 64, 112 \\
         JPEG 2000 & block width & 32, 64, 112 \\
         & tile heigth & 32, 64, 112 \\
         & tile width & 32, 64, 112 \\
         & progression order & \makecell{LRCP, RLCP,\\ PCLR, CPRL} \\
         \hdashline
         & quality & 0-100 \\
         & distance & 1.0-25.0 \\
         & effort & 1- 10 \\
         & progressive & - \\
         JPEG XL & compress boxes & 0, 1 \\
         & brotli effort & 0-11 \\
         & override bit depth & 8, 10, 12, 16 \\
         & resampling & -1, 1, 2, 4, 8 \\
         \hdashline
         JPEG AI & set target bpp & 1-50 \\
         \hdashline
         & qcolor & 0-100 \\
         & yuv & 444, 422, 420, 400 \\
         & sharp yuv & - \\
         & depth & 8, 10, 12 \\
         & premultiply & - \\
         AVIF & tilecolslog2 & 0-6 \\
         & tilerowslog2 & 0-6 \\
         & speed & 0-10 \\
         & range & full, limited \\
         & progressive & - \\
         \hdashline
         & quality & 0-100 \\
         & no alpha & - \\
         HEIF & bit depth & 8, 10, 12 \\
         & chroma downsampling & \makecell{average, nearest-neighbor,\\ sharp-yuv} \\
         & cut tiles & 32, 64, 112 \\
         \hdashline
         & preset & default, photo, picture \\
         & size & 1024 \\
         & method & 0-6 \\
         & segments & 1-4 \\
         & PSNR & 10-60 \\
         WebP & filter strength & 10-60 \\
         & sharpness & 0-7 \\
         & pass & 1-10 \\
         & SNS & 10-60 \\
         & no alpha & - \\
         \hline
    \end{tabular}
    \label{tab:compression_algorithm_settings}
\end{table}

\subsubsection{Color}
\label{subsubsec:color}
\begin{figure}[htbp]
    \centering    
    \begin{subfigure}[b]{0.2\textwidth}
        \centering
        \includegraphics[width=\textwidth]{gfx/preprocessing/original.png}
        \caption{}
        \label{fig:color}
    \end{subfigure}
    \hspace{0.2cm}
    \begin{subfigure}[b]{0.2\textwidth}
        \centering
        \includegraphics[width=\textwidth]{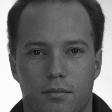}
        \caption{}
        \label{fig:gray}
    \end{subfigure}    
    \caption{This figure shows the impact of the grayscale conversion.}
    \label{fig:setup:grayscale_conversion}
\end{figure}

Every optimization is performed for color images and separately for grayscale images. The idea behind this step is that original images are stored with color, where each pixel is described by 24 bits, while after a grayscale conversion each pixel is only described by 8 bits, reducing the file size by a total of $\sim66\%$. Since this also results in a significant loss of information, all following steps are performed on both the color dataset as well as the grayscale dataset. The comparison of the images before and after conversion is shown in Fig. \ref{fig:setup:grayscale_conversion}.

\subsubsection{Resolution}
\label{subsubsec:resolution}
In their study, Grother et al.\@ \cite{grother_face_2025} have already shown that compression alone does not produce the best results, as it sometimes creates compression artifacts that ultimately distort faces. Instead, a combination of reduced resolution and image compression is better suited to reduce a facial image to small file sizes with as little loss of information as possible. The assumption here is as follows: The higher the resolution of an image, the more the compression algorithm has to compress. This inevitably leads to stronger compression artifacts. Reducing the resolution always deletes information but does not create any special compression artifacts. There should be an optimal ratio in which a compression algorithm causes less damage to the target image than a further reduction in resolution. The aim here is to find this ratio. Since running through all possible resolutions would be too computationally intensive, only rough steps in the reduction of resolutions are taken in the following. These include 112, which is the starting size because this resolution is required by AdaFace, as well as 56, 64, 75, 80, 96, 128, 160, 180, 200, and 224. Resolutions above 112 are used both directly from the original dataset and upscaled based on the required target size of 112, while resolutions below 112×112 pixels are created by downscaling the 112×112 images. FIQAT is always used for this adjustment. A special constraint for this step is given for JPEG AI, as this compression algorithm can only work with images with a resolution of 160×160 pixels or higher.

\subsubsection{Image manipulation}
\label{subsubsec:img_manipulation}
Although the individual compression algorithms differ in their implementation details, the overall procedure varies little, especially for the handcrafted algorithms. They all follow the same pattern, which, in simplified form, consists of the steps ``color space conversion'', ``image partitioning'', ``transform coding'', ``quantization'', and ``entropy encoding'' in order to reduce the file size. The fact that these overarching commonalities exist is by no means a coincidence, but stems from the general goal of image compression. The main focus is on keeping the quality losses of lossy compression that are perceptible to the human eye as small as possible. Various properties of human perception are exploited for this purpose. Particularly important here is the fact that humans perceive contrasts better than actual changes in color. Image regions with high contrast—i.e., edges—are therefore more important than regions of uniform color. For this reason, the compression algorithms are designed in such a way that they preserve more details in areas with high contrast, while in regions with lower contrast more information may be lost. \\
Since the extraction of feature vectors by AdaFace does not operate like human perception, it is possible that even minor visual changes caused by compression have a considerable impact on the extracted features. Therefore, the goal of image manipulation is to prepare the images in advance in a way that the more important regions, namely the area around mouth, nose, and eyes, of the image remain unchanged, while less important regions are modified so that the compression can achieve particularly large savings in these areas. \\
In order to achieve this goal, the first step must be to draw a clear distinction between these two areas, so that in a second step the less important areas can be manipulated accordingly. Two approaches are pursued for dividing the content into these two areas:
\begin{itemize}
    \item Rectangular approach: This method of differentiation is relatively simple and is shown visually in Fig. \ref{fig:rectangular_approach}. For this purpose, five landmarks are first detected using SCRFD \cite{scrfd}. These include the left and right eye, the tip of the nose, and the left and right corners of the mouth. Subsequently, from the image coordinates of these five points, the highest and lowest x and y coordinates are selected. Then, the smallest x and y coordinates are each reduced by 20\% of the total resolution, while the two largest values are each increased by 20\% of the total resolution. If a value falls below 0 or above the total resolution, it is capped at these values. As a result, a rectangle is spanned around the landmarks, whose interior is assumed to be the foreground and whose exterior is assumed to be the background.

    \begin{figure}[htbp]
    \centering    
    \begin{subfigure}[b]{0.15\textwidth}
        \centering
        \includegraphics[width=\textwidth]{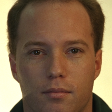}
        \caption{}
        \label{fig:rect_original}
    \end{subfigure}
    \hspace{0.2cm}
    \begin{subfigure}[b]{0.15\textwidth}
        \centering
        \includegraphics[width=\textwidth]{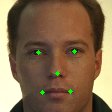}
        \caption{}
        \label{fig:rect_lm}
    \end{subfigure} \\ 
    \vspace{0.2cm}
    \begin{subfigure}[b]{0.15\textwidth}
        \centering
        \includegraphics[width=\textwidth]{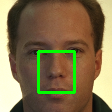}
        \caption{}
        \label{fig:rect_small_rect}
    \end{subfigure}
    \hspace{0.2cm}
    \begin{subfigure}[b]{0.15\textwidth}
        \centering
        \includegraphics[width=\textwidth]{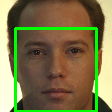}
        \caption{}
        \label{fig:rect_final_rect}
    \end{subfigure}    
    \caption{This figure shows the different stages of estimating the most important facial area. \ref{fig:rect_original} shows the original image. \ref{fig:rect_lm} shows the five landmarks detected by SCRFD \cite{scrfd}. \ref{fig:rect_small_rect} shows the rectangle including all landmarks. \ref{fig:rect_final_rect} shows the final estimation, by expanding the rectangle around the landmarks by 20\% in each direction.}
    \label{fig:rectangular_approach}
    \end{figure}
    
    \item OFIQ landmark-based approach: The second method to distinct between more and less important areas uses the OFIQ landmarks. These comprise 98 landmarks that outline the facial region around the mouth, nose, and eyes. In the subsequent process, the area inside these landmarks is labeled as most important. A depiction of this OFIQ landmark-based approach can be found in Fig. \ref{fig:ofiq_approach}. Note that this approach was added later on, so that manipulation operations that had already failed with the rectangular approach were no longer tested using this method.
    \begin{figure}[htbp]
    \centering    
    \begin{subfigure}[b]{0.15\textwidth}
        \centering
        \includegraphics[width=\textwidth]{gfx/preprocessing/50.png}
        \caption{}
        \label{fig:ofiq_original}
    \end{subfigure}
    \hspace{0.1cm}
    \begin{subfigure}[b]{0.15\textwidth}
        \centering
        \includegraphics[width=\textwidth]{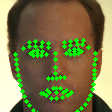}
        \caption{}
        \label{fig:ofiq_lm}
    \end{subfigure}
    \hspace{0.1cm}
    \begin{subfigure}[b]{0.15\textwidth}
        \centering
        \includegraphics[width=\textwidth]{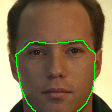}
        \caption{}
        \label{fig:ofiq_final}
    \end{subfigure}
    \caption{This figure illustrates the process of defining the most important facial area using OFIQ landmark detection. In \ref{fig:ofiq_original} the original image is shown. \ref{fig:ofiq_lm} shows the landmarks detected by OFIQ. \ref{fig:ofiq_final} shows the final approximation of the region of interest surrounding the detected landmarks.}
    \label{fig:ofiq_approach}
    \end{figure}
    \end{itemize}
    After separating the image regions, three types of modifications are made. These consist of:
    \begin{itemize}
    \item Unifying the less important area: In this process, all pixels that do not belong to the region around the mouth, nose, and eyes are replaced once with black pixels and once with white pixels. The respective result can be seen in Fig. \ref{fig:setup:unify_background}.
    \begin{figure}[htbp]
    \centering    
    \begin{subfigure}[b]{0.15\textwidth}
        \centering
        \includegraphics[width=\textwidth]{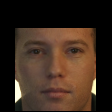}
        \caption{}
        \label{fig:blackout}
    \end{subfigure}
    \hspace{0.2cm}
    \begin{subfigure}[b]{0.15\textwidth}
        \centering
        \includegraphics[width=\textwidth]{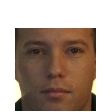}
        \caption{}
        \label{fig:whiteout}
    \end{subfigure}    
    \caption{This figure shows the impact of unifying the area outside the rectangular face estimation. In \ref{fig:blackout} all pixels have been replaced by black pixels, in \ref{fig:whiteout} the same was performed with white pixels.}
    \label{fig:setup:unify_background}
    \end{figure}
    
    \item Applying a 3×3 mean kernel: Instead of completely replacing the pixels, it can already be beneficial for compression if the pixels \textit{outside} the most important facial region become more similar. The major advantage over replacement is the preservation of information in the background, which is particularly useful in human-in-the-loop application scenarios. The application of the 3×3 mean kernel causes blurring, which reduces edges and increases the correlation between neighboring pixels. Both of these aspects can be advantageous for lossy compression. On the one hand, lossy compression algorithms are designed to invest more effort in preserving edges; on the other hand, they always use methods such as Wavelet Transform (WT) or the Discrete Cosine Transform (DCT), which exploit the correlation of neighboring pixels in natural photographs to reduce duplicated information. The different results of this operation can be seen in Fig. \ref{fig:blur}.
    \begin{figure}[htbp]
    \centering    
    \begin{subfigure}[b]{0.15\textwidth}
        \centering
        \includegraphics[width=\textwidth]{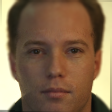}
        \caption{}
        \label{fig:rectangle_blur}
    \end{subfigure}
    \hspace{0.1cm}
    \begin{subfigure}[b]{0.15\textwidth}
        \centering
        \includegraphics[width=\textwidth]{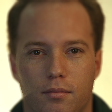}
        \caption{}
        \label{fig:ofiq_blur}
    \end{subfigure} 
    \hspace{0.1cm}
    \begin{subfigure}[b]{0.15\textwidth}
        \centering
        \includegraphics[width=\textwidth]{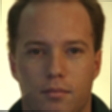}
        \caption{}
        \label{fig:full_blur}
    \end{subfigure}
    \hspace{0.2cm} 
    \caption{This figure includes the three applications of the 3×3 mean kernel. In \ref{fig:rectangle_blur} the blurred region is estimated using the rectangular approach, in \ref{fig:ofiq_blur} the OFIQ landmark-based approach is used, and in \ref{fig:full_blur} the entire image was smoothed. The latter operation logically also alters the more relevant regions of the face. However, since the blur does not change the basic geometry but only removes details, which also happens during the compression process, this variant is examined as well.}
    \label{fig:blur}
    \end{figure}    
    
    \item Low-pass filtering: The image is transformed into frequency domain using Fourier transformation. Then the upper 20\% of the frequencies are deleted. This operation works similar to a smoothing of the image and deletes minor details in texture while preserving the overall geometry of the image. Therefore, it could have similar effects to the previously presented 3×3 kernel variant and support lossy compression. Examples of the images after applying this image manipulation are shown in Fig. \ref{fig:lowpass}. Note that this type of image manipulation was initially implemented only for the grayscale images, since these only consist of one single channel. Since it already became apparent that it had a clearly negative effect on face recognition performance, we did not implement a variant for the color images.
    \begin{figure}[htbp]
    \centering    
    \begin{subfigure}[b]{0.15\textwidth}
        \centering
        \includegraphics[width=\textwidth]{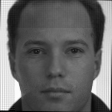}
        \caption{}
        \label{fig:rectangle_low}
    \end{subfigure}
    \hspace{0.2cm}
    \begin{subfigure}[b]{0.15\textwidth}
        \centering
        \includegraphics[width=\textwidth]{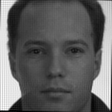}
        \caption{}
        \label{fig:full_low}
    \end{subfigure} 
    \caption{This figure shows the results after applying the low-pass filters to the grayscale images. This was implemented for the rectangular approach (\ref{fig:rectangle_low}) as well as a application to the full image (\ref{fig:full_low}), for the same reason the full smoothing variant was implemented.}
    \label{fig:lowpass}
    \end{figure}
\end{itemize}

\subsection{Parameter Optimization Process}
\label{subsec:params_optimize}
\begin{figure*}
    \centering
    \includegraphics[width=0.8\linewidth]{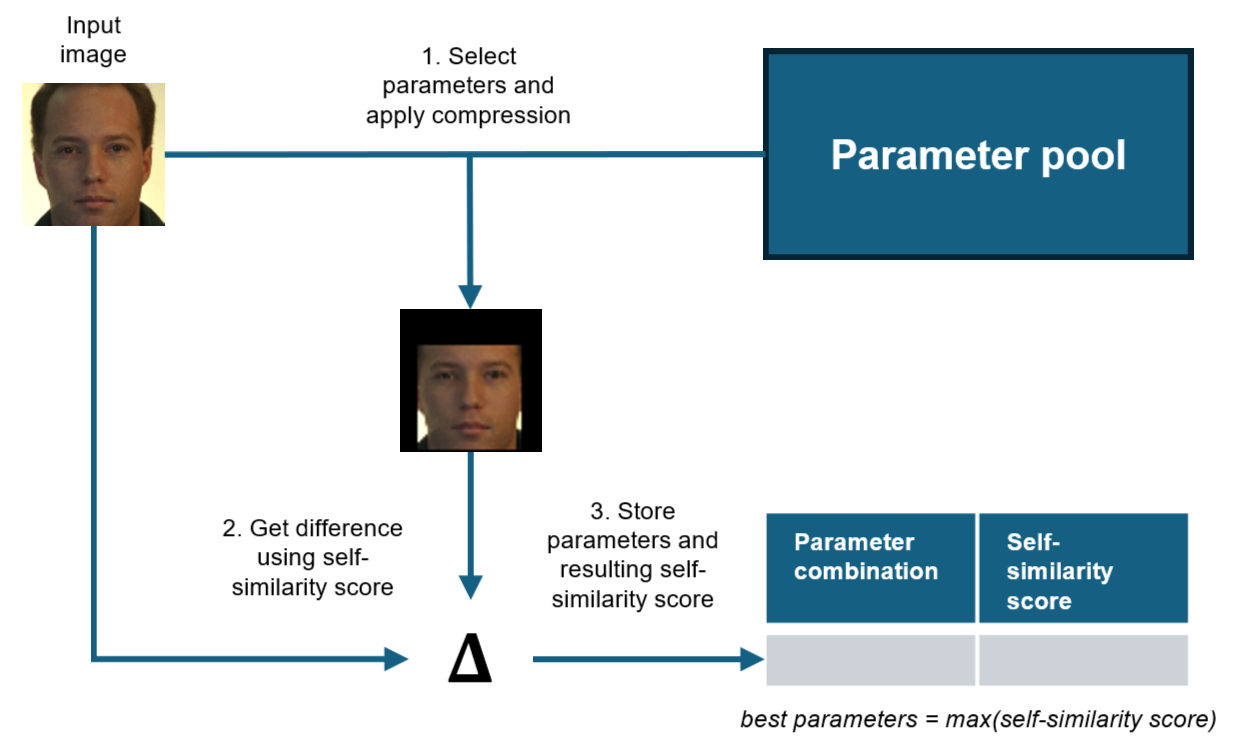}
    \caption{This figure shows the process implemented for evaluating the parameters in a simplified way.}
    \label{fig:params_optimize_proc}
\end{figure*}

The determination of the most suitable compression parameters takes place in a multi-stage process that can be divided into three phases. Initially, the influence of the parameters presented above is evaluated. The procedure chosen for this purpose is shown in simplified form in Fig. \ref{fig:params_optimize_proc}.
The core idea is to examine the effect of the selected parameters on the features extracted after compression. For this, a subset of the available parameters is applied to a facial image or to the compression process. Subsequently, AdaFace is used in order to extract a 512-dimensional feature vector from the original face image and the compressed version of $\leq1024$ bytes. Afterwards, the cosine similarity of the two resulting feature vectors is computed. The resulting scalar is the \textit{self-similarity score}, since it expresses the similarity between the original image and its compressed version.\\
This process is carried out for each parameter combination for all images in the parameter evaluation dataset. The average value of the calculated scalar is stored together with the parameters used. The selected parameter combinations are not selected randomly; instead, all parameters for the Command Line Interface (CLI) tools are first tested one by one. For each parameter, it is checked whether it has a positive, negative, or no effect on the resulting self-similarity score. For parameters that require a specific set of arguments, this is examined for each argument. For example, if the required argument is a number between 0 and 100, values are tested in steps of 10. For the best range, each value is then evaluated again in steps of 1. In a second step, all parameters without a positive effect are discarded. Finally, combinations of the remaining parameters are tested. The Python library Optuna \cite{optuna} is used for this purpose. It is designed for parameter optimization and offers the possibility of determining the best combination by performing a grid search on a defined set of parameters. For this case, the grid search is programmed to test the influence of each parameter on the resulting self-similarity score. \\
After the CLI tool parameters have been optimized, the best combination is selected to determine the most suitable resolution. This is again done using an Optuna grid search. \\
Afterwards, the best CLI parameters together with the best resolution are used to evaluate the various image manipulations. As a result, for each compression algorithm and each color variant, we obtain a combination of the parameters of the CLI tool, the resolution, and the image manipulation that produces the highest self-similarity score.\\
After this first phase is finished, for each combination of compression algorithm and color we select the three parameter combinations that produced the highest self-similarity scores. These are then applied to the frontal dataset. \\
For the following phases, we have slightly adapted the procedure compared to our preliminary study \cite{IWBF2026}. The idea was to better match the requirements of the application scenario of an ABC gate. The first difference lies in the extraction of the feature vectors from the original images. In \cite{IWBF2026}, the preprocessing consisting of cropping and aligning was first   applied to every image of the full dataset. As a result, all images were available at a resolution of 112×112, but the features had not yet been extracted in this step. The frontal images were then manipulated and compressed accordingly. The compressed images were subsequently realigned and cropped, i.e., again resized to a size of 112×112 pixels. The uncompressed probe images used for comparison were already available at the correct resolution at this point; however, they too were realigned and cropped again before feature extraction. This approach was chosen so that the compressed images and the original images would be processed as similar as possible. Here, the second alignment of the uncompressed facial images often ended up being slightly different, so they were also resized a second time. We have revised this procedure for the present version, because in the practical application of an ABC gate this second processing of the uncompressed images is not required and would therefore unnecessarily lengthen the identification process, which is undesirable. A second important change compared to \cite{IWBF2026} is the determination of the concrete threshold values used for labeling similarity scores as matches and non-matches. While we previously used linear interpolation to determine these values, in this work we use the lowest possible values. This approach corresponds better to the operational scenario and was chosen for both scenarios in order to prevent methodological differences. However, both of these changes result in slightly deviates for some cases compared to the values reported in \cite{IWBF2026}. \\
In the second phase the face recognition performance of the resulting datasets is evaluated by comparing every compressed image against all images of the full dataset. In this case, the threshold is set to a FMR of 0.01\%. \\ 
In the third and final phase, each compressed image is then compared with every image from the frontal dataset. This comparison is carried out because it closely matches the application scenario at the ABC gate, where no strong pose variations must be expected. In this phase, the threshold is set to an FMR of 0.0001\%, which can better reveal performance differences in terms of FNMR. At the end of the evaluation process, the most suitable compression algorithm with optimized parameters can then be determined based on the best face recognition performance.

\section{Results}
\label{sec:results}
This section deals with the respective results of the optimization process presented above. The optimized parameters that were found are presented, the changes in the similarity scores obtained from the feature vectors of the images are described, and the resulting effects on face recognition performance are analyzed. For this purpose, both the different FNMR and the respective similarity scores are examined.

\subsection{Optimized Parameters}
\label{subsec:optimized_params}
With regard to the optimized parameters, a distinction must be made between the parameters most suitable for comparison with the full dataset and those most suitable for comparison with the frontal dataset. 

\subsubsection{Full Comparison}
\label{subsubsec:full_comparison}
In the following, the results of the comparison against the full dataset are discussed first. The results can be found in Table \ref{tab:results:parameters_full}. 
\begin{table*}[!htbp]
\centering
\caption{Final optimal parameter sets for comparing compressed images against the full dataset}
\begin{tabular}{lcccc}
\hline
\textbf{\makecell{Compression \\ algorithm}} & 
\textbf{\makecell{CLI arguments \\ and values}} & 
\textbf{\makecell{Color}} & 
\textbf{\makecell{Resolution}} & 
\textbf{\makecell{Image \\ manipulation}} \\
\hline
\textbf{JPEG}      & \makecell{grayscale \\ arithmetic \\ smooth = 30} & Grayscale & 96×96 & rectangular blur \\
\hdashline
\textbf{JPEG 2000} & \makecell{ratio \\ number of resolutions = 3} & Color & 56×56 & rectangular blur \\
\hdashline
\textbf{JPEG XL}   & \makecell{quality \\ effort = 10} & Color & 64×64 & rectangular blur \\
\hdashline
\textbf{JPEG AI}   & \makecell{n.a.} & Color & 180×180 & \makecell{OFIQ landmark-based blur} \\
\hdashline
\textbf{AVIF}      & \makecell{yuv = 420 \\ speed = 1} & Color & 56×56 & rectangular blur \\
\hdashline
\textbf{HEIF}      & \makecell{chroma downsampling = average} & Color & 96×96 & rectangular blur \\
\hdashline
\textbf{WebP}      & \makecell{method = 6 \\ sns = 40} & Color & 64×64 & rectangular blur\\
\hline
\end{tabular}
\label{tab:results:parameters_full}
\end{table*}
Some commonalities between the compression algorithms can be seen from Table \ref{tab:results:parameters_full}. Looking at the “Color” column, it becomes apparent that for this scenario choosing color images is the better option for almost all algorithms. The only exception here is JPEG. This compression algorithm frequently fails to achieve the specified target size of $\leq1024$ bytes for color images. For this reason, grayscale images must be used in this case, as they are initially already about 60\% smaller. \\
With respect to resolution, however, it emerges that it should generally be chosen lower than the 112×112 pixels required for face recognition, which is in line with the findings of Grother et al. \cite{grother_face_2025}. The optimal resolution, however, varies from algorithm to algorithm. Here, JPEG AI is the only exception due to the already mentioned property that this compression algorithm currently supports only resolutions of 160×160 pixels and higher. \\
A clear trend is also evident in the “Image manipulation” column. Applying the 3×3 mean kernel has a positive impact on the overall result in all cases. Interestingly, the simpler rectangular blur approach outperforms the more precise OFIQ landmark-based blur in all cases except for JPEG AI.

\subsubsection{Frontal Comparison}
\label{subsubsec:frontal_comparison}
After the results from phase two of the optimization process are now known, the results of the third phase are presented. Here, the compressed face images are compared only against the frontal dataset, that is, against face images of high face image quality with little or no pose variation. The result can be found in Table \ref{tab:results:parameters_frontal}.
\begin{table*}[!htbp]
\centering
\caption{Final optimal parameter sets for comparing compressed images against frontal dataset}
\begin{tabular}{lcccc}
\hline
\textbf{\makecell{Compression \\ algorithm}} & 
\textbf{\makecell{CLI arguments \\ and values}} & 
\textbf{\makecell{Color}} & 
\textbf{\makecell{Resolution}} & 
\textbf{\makecell{Image \\ manipulation}} \\
\hline
\textbf{JPEG}      & \makecell{grayscale \\ arithmetic \\ smooth = 30} & Grayscale & 96×96 & rectangular blur \\
\hdashline
\textbf{JPEG 2000} & \makecell{ratio \\ number of resolutions = 5} & Grayscale & 56×56 & rectangular blur \\
\hdashline
\textbf{JPEG XL}   & \makecell{quality \\ effort = 10} & Grayscale & 56×56 & \makecell{OFIQ landmark-based blur} \\
\hdashline
\textbf{JPEG AI}   & \makecell{n.a.} & Grayscale & 200×200 & full blur \\
\hdashline
\textbf{AVIF}      & \makecell{YUV format = 400 \\ speed = 1} & Grayscale & 56×56 & no manipulation \\
\hdashline
\textbf{HEIF}      & \makecell{chroma downsampling = average} & Color & 96×96 & rectangular blur \\
\hdashline
\textbf{WebP}      & \makecell{method = 6 \\ sns = 40} & Grayscale & 80×80 & \makecell{OFIQ landmark-based blur} \\
\hline
\end{tabular}
\label{tab:results:parameters_frontal}
\end{table*}
Some commonalities can also be observed in Table \ref{tab:results:parameters_frontal}. For this configuration, grayscale images prove to be the more suitable choice. The only exception here is HEIF. Once again, it becomes apparent that, in general, a resolution below 112×112 delivers better results for all compression algorithms except JPEG AI. A familiar pattern also emerges for image manipulation: Smoothing is once again the preferred strategy. The only exception is now AVIF; for this compression algorithm, the images should not be further modified. In addition, the choice of the region to be smoothed now varies more strongly than in the previous setup. For JPEG AI, it now turns out that smoothing the entire facial image with the 3×3 kernel is the most suitable approach, without distinguishing between more and less important regions.

\subsection{Compressed Images}
\label{subsec:compr_img}
This section presents several compressed images in order to gain a better understanding of the effect of strong lossy compression and various parameters on the resulting images.
\begin{figure}[htbp]
\centering    
\begin{subfigure}[b]{0.22\textwidth}
    \centering
    \includegraphics[width=\textwidth]{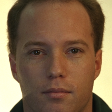}
    \caption{Original}
    \label{fig:compr_img:color:original}
\end{subfigure}
\hspace{0.2cm}
\begin{subfigure}[b]{0.22\textwidth}
    \centering
    \includegraphics[width=\textwidth]{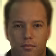}
    \caption{AVIF}
    \label{fig:compr_img:color:avif}
\end{subfigure} \\
\vspace{0.2cm}
\begin{subfigure}[b]{0.22\textwidth}
    \centering
    \includegraphics[width=\textwidth]{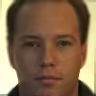}
    \caption{HEIF}
    \label{fig:compr_img:color:heif}
\end{subfigure} 
\hspace{0.2cm}
\begin{subfigure}[b]{0.22\textwidth}
    \centering
    \includegraphics[width=\textwidth]{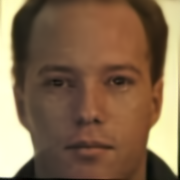}
    \caption{JPEG AI}
    \label{fig:compr_img:color:jai}
\end{subfigure} 
\caption{This figure shows the results, when using optimized parameters for compressing face images for comparison against images of unknown face image quality. Note: The images have all been scaled to the same size for improved viewing. As a result, the differences in proportions due to the varying resolutions are no longer visible.}
\label{fig:compr_img:color}
\end{figure}
\begin{figure}[htbp]
\centering    
\begin{subfigure}[b]{0.22\textwidth}
    \centering
    \includegraphics[width=\textwidth]{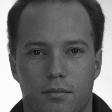}
    \caption{Original}
    \label{fig:compr_img:gray:original}
\end{subfigure}
\hspace{0.2cm}
\begin{subfigure}[b]{0.22\textwidth}
    \centering
    \includegraphics[width=\textwidth]{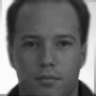}
    \caption{JPEG}
    \label{fig:compr_img:gray:jpeg}
\end{subfigure} \\
\vspace{0.2cm}
\begin{subfigure}[b]{0.22\textwidth}
    \centering
    \includegraphics[width=\textwidth]{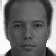}
    \caption{JPEG XL}
    \label{fig:compr_img:gray:jxl}
\end{subfigure} 
\hspace{0.2cm}
\begin{subfigure}[b]{0.22\textwidth}
    \centering
    \includegraphics[width=\textwidth]{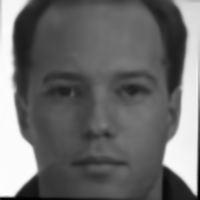}
    \caption{JPEG AI}
    \label{fig:compr_img:gray:jai}
\end{subfigure} 
\caption{This figure shows the results, when using optimized parameters for compressing face images for comparison against images of high face image quality (ABC gate scenario). Note: The images have all been scaled to the same size for improved viewing. As a result, the differences in proportions due to the varying resolutions are no longer visible.}
\label{fig:compr_img:gray}
\end{figure}
Both in Fig. \ref{fig:compr_img:color} and in Fig. \ref{fig:compr_img:gray} it can be seen that, due to lossy compression, the detailed skin texture is lost. It is also apparent that as the resolution decreases, the textures in general become increasingly coarse. For both the color images and the grayscale images, it is evident that the details in the eyes are noticeably degraded, especially the white areas next to the iris are affected by this. Since JPEG AI operates at a significantly higher resolution than the other compression algorithms in both scenarios, these images appear particularly well preserved by comparison for the human perception.

\subsection{Self-similarity Scores}
\label{subsec:sss}
The expectation for the self-similarity scores is that they express the deviations of a compressed facial image from those of the original in a single scalar value. For this reason, we used this metric for parameter optimization in the first phase. Here only the parameter search dataset was used. Table \ref{tab:results:sss} includes the averaged scores for the parameter search dataset as well as the frontal dataset.
\begin{table}[!htbp]
\setlength{\tabcolsep}{4pt}
\centering
\caption{Averaged self-similarity scores for the final settings}
\renewcommand{\arraystretch}{1.2}
\begin{tabular}{lcccc}
\toprule
\textbf{\makecell{\\Compression \\ algorithm}} 
& \multicolumn{2}{c}{\textbf{Full Comparison}} 
& \multicolumn{2}{c}{\textbf{Frontal Comparison}} \\
\cmidrule(lr){2-3}
\cmidrule(lr){4-5}
& \textbf{\makecell{Parameter \\ search \\ dataset \\ $\uparrow$}} 
& \textbf{\makecell{Frontal \\ dataset \\ $\uparrow$}} 
& \textbf{\makecell{Parameter \\ search \\ dataset \\ $\uparrow$}} 
& \textbf{\makecell{Frontal \\ dataset \\ $\uparrow$}} \\
\midrule
\textbf{JPEG}      & 0.9560 & 0.9190 & 0.9560 & 0.9199 \\
\textbf{JPEG AI}   & 0.9217 & 0.8938 & 0.9689 & 0.9484 \\
\textbf{AVIF}      & 0.9076 & 0.8862 & 0.9348 & 0.9173 \\
\textbf{WebP}      & 0.9040 & 0.8902 & 0.9383 & 0.9212 \\
\textbf{HEIF}      & 0.8789 & 0.8596 & 0.8789 & 0.8596 \\
\textbf{JPEG XL}   & 0.8735 & 0.8597 & 0.9042 & 0.8648 \\
\textbf{JPEG 2000} & 0.8502 & 0.8553 & 0.9065 & 0.8601 \\
\bottomrule
\end{tabular}
\label{tab:results:sss}
\end{table}
It shows an interesting tendency. The self-similarity score for grayscale facial images is always higher than the score for the color images. This explains why JPEG reaches by far the highest score for the full comparison, since this algorithm always uses grayscale images. It is exactly the other way around with HEIF, where color images are always used, which in the frontal comparison on the parameter search dataset leads to a lower value than for all the other algorithms. \\
Additionally, it also becomes apparent that the compression algorithms JPEG, JPEG AI, AVIF, and WebP produce higher self-similarity scores than the other three algorithms. In this context, JPEG is a surprising candidate, as this algorithm has repeatedly proven to be poorly suited in previous studies and is known for its blocking artefacts at high compression ratios.\\
Another positive finding from Table \ref{tab:results:sss} is that the optimizations based on the parameter search dataset apparently transfer to the significantly larger frontal dataset. Given the possible value range from -1 to 1, the differences between the two datasets are no greater than 2.5\%.

\subsection{Face Recognition Performances}
\label{subsec:fr_performances}
The metrics addressed in this section are the FMR and the FNMR. These indicate how many errors a face recognition system would make when operating on the data. Here, much depends on the selected threshold. This determines what ratio of false matches and false non-matches occurs. For this purpose FRONTEX, the European Border and Coast Guard Agency, defines the following guideline: a given system, at a FMR of 0.1\%, should not exceed a FNMR of 5.0\% \cite{european_agency_for_the_management_of_operational_cooperation_at_the_external_borders_of_the_member_states_of_the_european_union_guidelines_2016}. Since for all compression algorithms this FMR threshold of 0.1\% would already result in no false non-match errors, which would complicate the comparison of the algorithms, we select thresholds for each scenario that meet this minimum requirement while still allowing a good ranking of the algorithms. As before, full comparison and frontal comparison are treated separately.

\subsubsection{Face Recognition Performance for Full Comparison}
\label{subsubsec:fr_full_comparison}
For this investigation, the 2,638 compressed frontal facial images are compared against each of the 11,128 images of the full dataset. Since face recognition fails in some cases due to the poor face image quality of certain images in the full dataset, this results in 2,638 self-similarity scores, 39,857 mated similarity scores, and 29,301,902 non-mated similarity scores. We set the threshold to a FMR of 0.01\%, i.e., the largest possible value after 2,932 false matches. This threshold is determined individually for each algorithm. The overall result of these data can be seen in Table \ref{tab:results:fr_full}.
\begin{table}[!htbp]
\setlength{\tabcolsep}{4pt}
\centering
\caption{Face recognition performance of each compression algorithm at a false match rate of 0.01\%, when comparing the compressed frontal dataset against the full dataset}
\renewcommand{\arraystretch}{1.2}
\begin{tabular}{lccc}
\toprule
\textbf{\makecell{Compression \\ algorithm}} 
& \textbf{\makecell{FNMR in \% \\ at FMR = 0.01\% \\ $\downarrow$}} 
& \textbf{Threshold} 
& \textbf{\makecell{False \\ non-matches \\ in total \\ $\downarrow$}} \\
\midrule
\textbf{\makecell{no compression \\ (baseline)}} & 0.0100 & 0.2566 & 4 \\
\hdashline
\textbf{WebP}       & 0.0151 & 0.2528 & 6 \\
\textbf{AVIF}       & 0.0151 & 0.2523 & 6 \\
\textbf{JPEG AI}    & 0.0151 & 0.2493 & 6 \\
\textbf{JPEG XL}    & 0.0176 & 0.2531 & 7 \\
\textbf{JPEG 2000}  & 0.0251 & 0.2550 & 10 \\
\textbf{HEIF}       & 0.0276 & 0.2508 & 11 \\
\textbf{JPEG}       & 0.0853 & 0.2541 & 34 \\
\bottomrule
\end{tabular}
\label{tab:results:fr_full}
\end{table}
As expected, Table \ref{tab:results:fr_full} shows that compressing the face images generally has a negative effect on face recognition performance. Nevertheless, all values lie within the range specified by FRONTEX. The most suitable formats here are WebP, AVIF, and JPEG AI. Here, the impact of the changed choice of threshold compared to our previous work \cite{IWBF2026} becomes clear. The FNMR of WebP, JPEG AI, JPEG XL, and JPEG has each improved slightly. This also results in the different ordering as well as the selection of the most suitable compression algorithm(s). Despite its high self-similarity scores, JPEG delivers by far the worst result. This is mainly due to the use of grayscale images. Apart from that, the tendencies of the compression algorithms in terms of face recognition performance coincide with those of the self-similarity scores.

\subsubsection{Face Recognition Performance for Frontal Comparison}
\label{subsubsec:fr_frontal_comparison}
When comparing the compressed face images with those of the frontal dataset, 2,638 self-similarity scores, 6,968 mated similarity scores, and 6,949,438 non-mated similarity scores are obtained. Since this now involves a comparison against images of very high face image quality, there are no images where face recognition fails. In addition, the threshold must be set significantly higher in order to get any error rates. We therefore choose an FMR of 0.0001\%, which in this case results in a total of 8 false matches. The results for this are shown in Table \ref{tab:results:fr_frontal}.
\begin{table}[!htbp]
\setlength{\tabcolsep}{4pt}
\centering
\caption{Face recognition performance of each compression algorithm at a false match rate of 0.0001\%, when comparing the compressed frontal dataset against the frontal dataset}
\renewcommand{\arraystretch}{1.2}
\begin{tabular}{lccc}
\toprule
\textbf{\makecell{Compression \\ algorithm}} 
& \textbf{\makecell{FNMR in \% \\ at FMR = 0.0001\% \\ $\downarrow$}} 
& \textbf{Threshold} 
& \textbf{\makecell{False \\ non-matches \\ in total \\ $\downarrow$}} \\
\midrule
\textbf{\makecell{no compression \\ (baseline)}} & 0.4592 & 0.5864 & 32 \\
\hdashline
\textbf{JPEG AI}    & 0.0718 & 0.5173 & 5 \\
\textbf{AVIF}       & 0.2870 & 0.5238 & 20 \\
\textbf{JPEG}       & 0.3157 & 0.5166 & 22 \\
\textbf{WebP}       & 0.4162 & 0.5329 & 29 \\
\textbf{HEIF}       & 0.7032 & 0.5237 & 49 \\
\textbf{JPEG 2000}  & 0.7319 & 0.4959 & 51 \\
\textbf{JPEG XL}    & 0.8037 & 0.5041 & 56 \\
\bottomrule
\end{tabular}
\label{tab:results:fr_frontal}
\end{table}
This table now shows some changes compared to the previous results. The fact that the FNMR has increased significantly is due to the fact that the FMR is, in return, considerably lower. What immediately stands out is that, contrary to expectations, compression now has a positive effect on face recognition performance in some cases. JPEG AI clearly outperforms the other algorithms this time. Another unexpected development is the good performance of JPEG. While this compression algorithm previously delivered by far the worst results, it now even surpasses the performance of the original images. Apart from that, however, the distribution of the algorithms remains similar. \\
The fact that, in this scenario, the performance after compression in the case of JPEG AI, AVIF, JPEG, and WebP has increased compared to that of the original images can be explained by the noise-filtering property of the compression, as described by Bousnina et al.\@ \cite{deep_learning_impact}. This reduces the noise present in the original images, which can counteract the performance loss caused by the deletion of information.

\subsection{Similarity Score Distributions}
\label{subsec:ssd}
After the previous section introduced the changes in face recognition performance, this section examines in detail the effects of compression on the mated and non-mated similarity scores. The focus is not only on the impact of the two datasets used for comparison, but also on the difference caused by the choice between color and grayscale images. Since the distributions of the respective compression algorithms are very similar in the overall picture, in the following only the distributions of the comparisons of the JPEG AI-compressed face images will be examined as representative.

\subsubsection{Similarity Score Distributions for Full Comparison}
\label{subsubsec:ssd_full} 
\begin{figure*}
    \centering
    \includegraphics[width=0.8\textwidth]{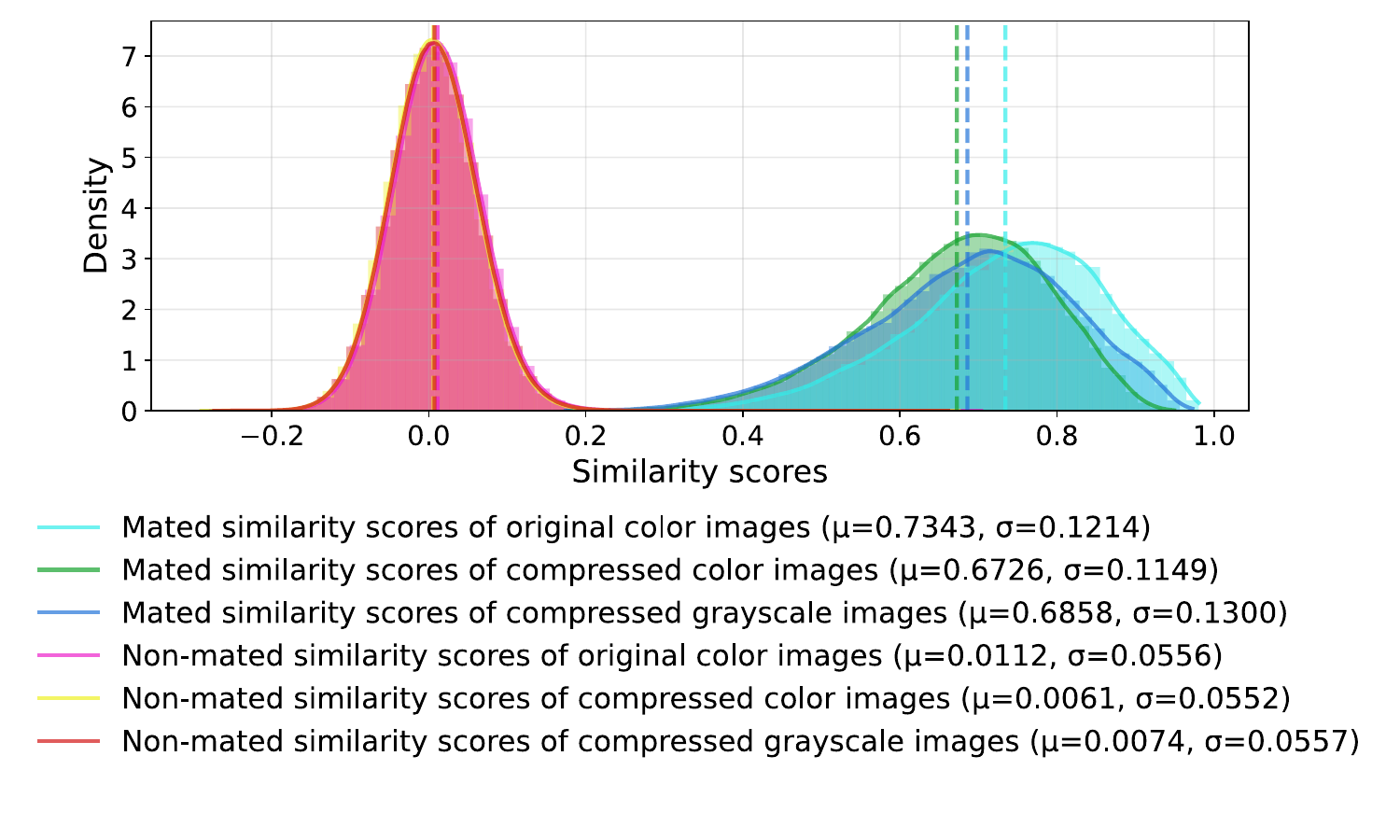}
    \caption{This figure shows the mated and non-mated similarity score distributions obtained by comparing all images included in the frontal dataset (original, color JPEG AI-compressed, and grayscale JPEG AI-compressed) against all images included in the full dataset (original, corresponding colors).}
    \label{fig:ssd_full}
\end{figure*}

Fig. \ref{fig:ssd_full} shows the distributions of the mated and non‑mated similarity scores for the comparison of the original frontal dataset with the full dataset, the JPEG AI‑compressed color frontal dataset with the color full dataset, and the JPEG AI‑compressed grayscale frontal dataset with the grayscale full dataset.
The following observations can be made: The impact of compression with the different settings on the non-mated scores is comparatively small. Neither the overall distribution of these scores nor the mean value changes significantly. This shows that dissimilar faces do not become more dissimilar due to compression. However, a small but very important development can be observed in the tails of the non-mated similarity score distributions. Here it becomes apparent that the ends of the distributions contract due to compression, which means that the particularly similar non-mated faces become less similar through compression. This effect is particularly strong for the grayscale images, but can also be clearly observed for the compressed color images. This effect alone would lead to improved face recognition performance, which, however, was not observed as shown in Section \ref{subsubsec:fr_full_comparison}.\\
The situation is different for the development of the mated similarity score distributions. These are significantly reduced by compression, which can be clearly seen from the mean values. This development explains why face recognition performance decreased overall despite the improvement of the non-mated similarity scores. An interesting detail here is that the compressed grayscale images produce, on average, higher mated similarity scores than the color images. The fact that the color images nevertheless yield better face recognition performance in this scenario is due to the higher standard deviation of the distribution of the grayscale images.

\subsubsection{Similarity Score Distributions for Frontal Comparison}
\label{subsubsec:ssd_frontal}
\begin{figure*}
    \centering
    \includegraphics[width=0.8\textwidth]{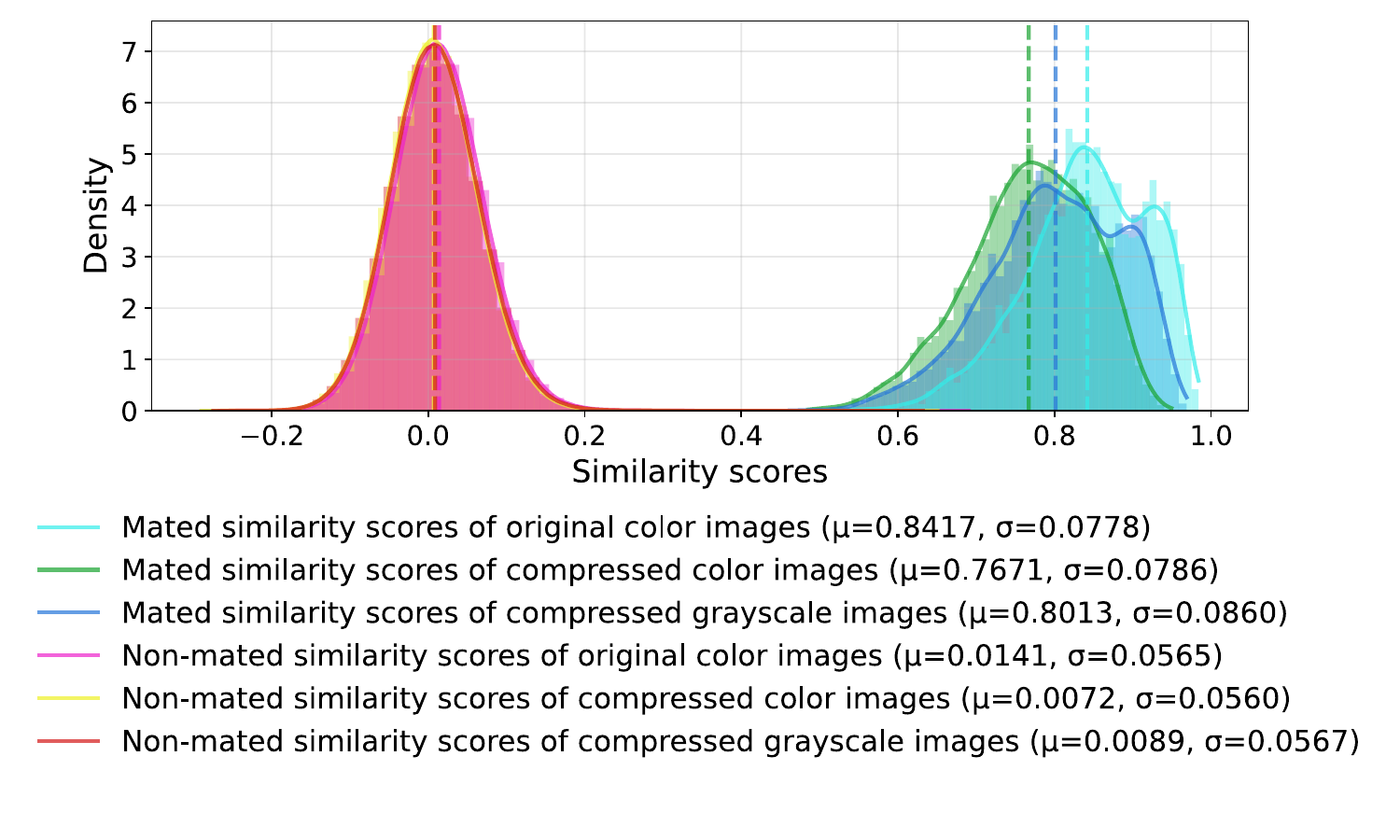}
    \caption{This figure shows the mated and non-mated similarity score distributions obtained by comparing all images included in the frontal dataset (original, color JPEG AI-compressed, and grayscale JPEG AI-compressed) against all images included in the frontal dataset (original, corresponding colors).}
    \label{fig:ssd_frontal}
\end{figure*}

In Fig. \ref{fig:ssd_frontal}, the mated and non-mated similarity score distributions are shown for the comparisons of the original frontal dataset, the color JPEG AI-compressed frontal dataset, and the grayscale JPEG AI-compressed frontal dataset with the respective frontal dataset of the same color.
The developments already observed in the investigation with the full dataset can be seen again. Once more, compression has little effect on the non-mated similarity score distributions, with the exception of the right tail, where the highest values have again decreased in the same way. The standard deviations have also not changed noticeably.\\
The previously observed trend can also be seen again for the mated similarity score distributions. This time, however, the difference between the distribution of the original images and that of the JPEG AI-compressed grayscale images is significantly smaller than before. In contrast, the difference between the compressed grayscale images and the compressed color images is considerably larger. The standard deviations of the mated similarity score distributions resulting from the frontal comparison are significantly smaller than those from the full comparison, which is due to the exclusive use of images with high face image quality. In the general course of the mated similarity score distributions shown in Fig. \ref{fig:ssd_frontal}, it can also be seen that the distribution of the grayscale images corresponds to  that of the original images much more closely. Since the reduction in the mated similarity scores of the grayscale images is much smaller here, while the decrease in the extremely high non-mated similarity scores remains the same, this also explains why the face recognition performance of the grayscale images not only surpasses that of the color images, but in some cases even that of the original images. The storage capacity freed up by omitting the color information, and the resulting better-preserved contrast details, appear in this scenario to outweigh the loss of color information.

\subsection{Face Image Quality Assessment}
\label{subsec:fiqa}
\begin{figure}
    \centering
    \includegraphics[width=\linewidth]{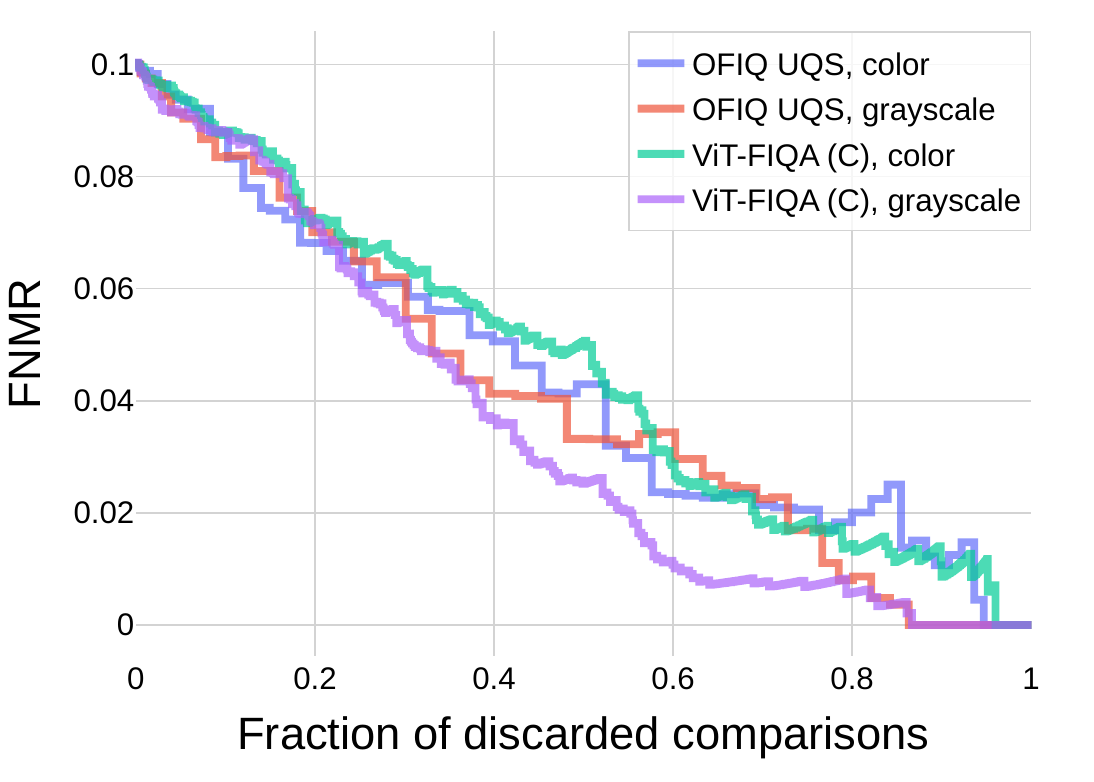}
    \caption{This figure shows the EDC curves for both JPEG AI-compressed color images and JPEG AI-compressed grayscale images using OFIQ UQS and ViT-FIQA(C) for FIQA.}
    \label{fig:edc}
\end{figure}

In addition to the previous investigations, we now examine to what extent face image quality assessment (FIQA) algorithms can handle the highly compressed images. For this purpose, based on the compressed frontal dataset, we calculate Error versus Discard Characteristic (EDC) curves using the FIQA algorithms OFIQ UQS and ViT-FIQA(C). As in the previous sections, we again discuss only one exemplary EDC plot, namely the plot of JPEG AI. This is shown in Fig. \ref{fig:edc}.
The EDC curves all show decreasing false non-match rates and indicate that the FIQA algorithms are able to handle the compressed images. It is also evident that ViT-FIQA(C) can assess the face image quality of the compressed grayscale images better than of the color variants. While ViT-FIQA(C) provides particularly good assessments for the grayscale images, the OFIQ UQS performance for the color and grayscale variants appears less differentiated. Overall, it performs worse than ViT-FIQA(C) for grayscale images, but better for color images. With a fraction of discarded comparisons of around 60\%, the EDC curve of ViT-FIQA(C) for the grayscale images begins to noticeably flatten out. Since a system that discards so many samples is of no practical use, this does however seem less relevant.

\section{Conclusion}
\label{sec:conclusion}
The results of our investigations showed that a strong reduction in file size is possible without degrading face recognition performance to the point where a corresponding system becomes useless. In particular, the JPEG AI compression algorithm proved to be a suitable candidate in both scenarios, as it not only provides a very good compression rate, but at the same time maintains a comparatively high resolution, which presumably further supports human-in-the-loop scenarios. However, the compression algorithms AVIF and WebP were also able to deliver competitive results in both considered scenarios. \\
This recommendation differs from our previous findings \cite{IWBF2026}, because JPEG AI has shown superior performance for the now explicitly considered application scenario of an ABC gate. At the same time, its performance in the general scenario has improved to that of AVIF thanks to the adjustments described in \ref{subsec:params_optimize}, which in turn enables a uniform recommendation. \\
The results also showed that reducing the resolution and blurring the images before applying lossy compression has a positive impact on the face recognition performance of the outcomes, which again is in line with prior literature \cite{grother_face_2025}\cite{deep_learning_impact}\cite{finger_vein_compression}. \\
A small surprise is provided by JPEG, the oldest compression algorithm considered here. After appropriate parameter optimization, it was able to deliver results for grayscale images that could keep up with the top candidates in the more restricted ABC gate scenario. The HEIF and JPEG 2000 algorithm, on the other hand, should not be used in contexts with strong lossy compression. \\
Less surprising is the finding that comparison of exclusively frontal images, even when using heavily compressed facial images, leads to significantly better results with no loss of face recognition performance in contrast to a comparison against images without biometrically optimal conditions. It was shown that grayscale conversion is a well-suited measure to reduce file size when comparing images with higher face image quality, whereas for comparison against images of lower quality it leads to significantly worse results. \\
Both FIQA algorithms used are able to handle the compressed images. In this case, ViT-FIQA(C) is better suited for the grayscale images, whereas OFIQ UQS provides better results for the color images.

\section{Future Work}
\label{sec:future_work}
This paper shows that a state-of-the-art face recognition system remains functional even with heavily compressed images. In particular, for the ABC gate application scenario, it becomes apparent that, with an appropriate choice of parameters, face images of 1024 bytes are sufficient to achieve face recognition performance that is only slightly, if at all, degraded. Nevertheless, future studies should investigate additional factors. For example, in this work only AdaFace was used for feature extraction. This algorithm is state-of-the-art and openly available. A more in-depth investigation should also take the wider range  of commercial systems into consideration. \\
One aspect that was neglected here is the encoding and decoding times of the individual compression algorithms. For the ABC gate application scenario, these are only of secondary importance, since the main time cost arises from encoding during the enrolment phase, i.e., during the document issuing process. As only decoding occurs during the identification process at the gate, this metric was not of interest for our study. For other applications, however, it could be more important, so additional investigations in this direction would be useful in the future.

\section{Acknowledgment}
Portions of the research in this paper use the FERET database of facial images collected under the FERET program, sponsored by the DOD Counterdrug Technology Development Program Office.

\bibliographystyle{IEEEtran}
\bibliography{bibliography}

\vfill\break

\begin{IEEEbiography}[{\includegraphics[width=1in,height=1.25in,clip,keepaspectratio]{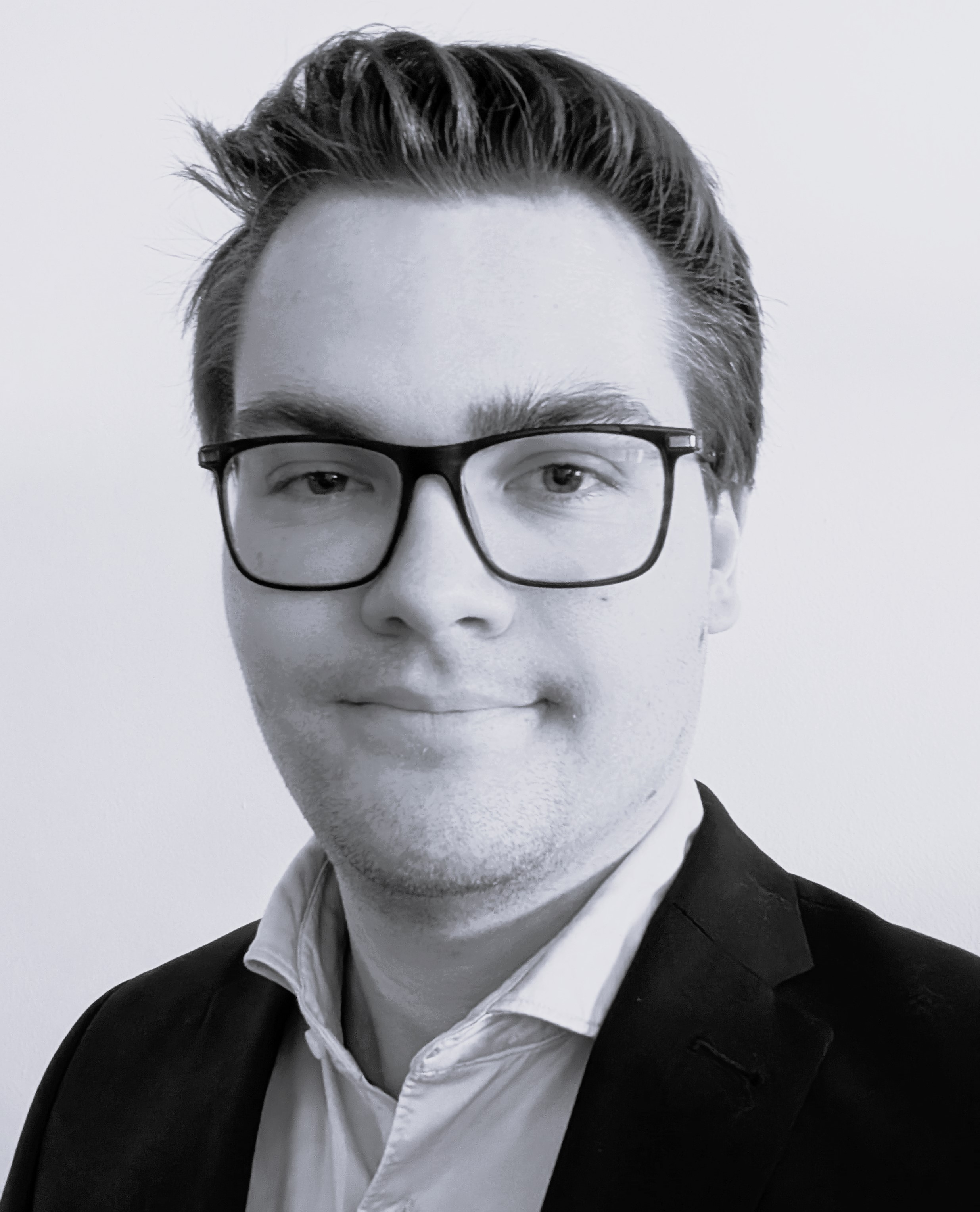}}]{Paul Andreas} received the B.S. degree in information science in 2024 and the M.S. degree in information science from Hochschule Darmstadt (HDA), Darmstadt, Germany, in 2026.
\end{IEEEbiography}

\begin{IEEEbiography}[{\includegraphics[width=1in,height=1.25in,clip,keepaspectratio]{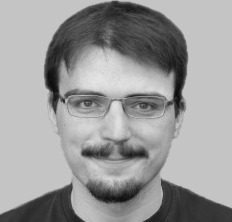}}]{Torsten Schlett} received the M.Sc. degree in computer science from Hochschule Darmstadt (HDA), Germany, in 2019. He is currently pursuing the Ph.D. degree with the da/sec Biometrics and Security Research Group, which is affiliated with HDA and the National Research Center for Applied Cybersecurity (ATHENE). He has been involved in the EU project iMARS and standardization activities of ISO/IEC JTC1 SC37. His main research area is face image quality assessment.
\end{IEEEbiography}

\begin{IEEEbiography}[{\includegraphics[width=1in,height=1.25in,clip,keepaspectratio]{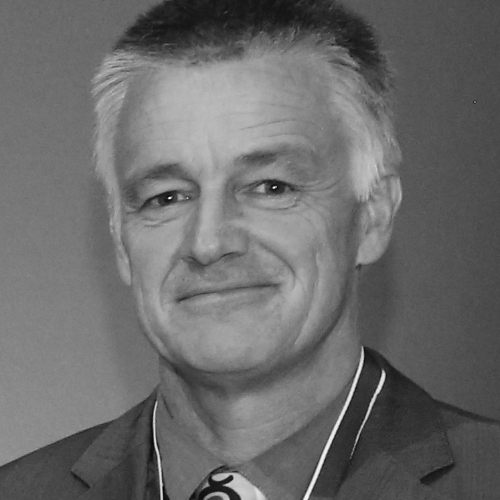}}]{Christoph Busch} (Fellow, IEEE) is member of NTNU-Gjøvik, Norway and holds a joint appointment with Hochschule Darmstadt (HDA), Germany. On behalf of the German BSI he has been the coordinator for the project series BioIS, BioFace, BioFinger, BioKeyS Pilot-DB, KBEinweg and NFIQ2.0.  He was initiator and participated in multiple European projects on biometrics (e.g.\@ 3D-Face, FIDELITY and iMARS). He is also PI in ATHENE. He is co-founder of the European Association for Biometrics (EAB). He co-authored more than 700 technical papers and has been a speaker at international conferences. He is member of the editorial board of the IET journal on Biometrics and formerly of the IEEE TIFS journal. Furthermore he chairs the TeleTrusT biometrics working group as well as the German standardization body on Biometrics and is convenor of WG3 in ISO/IEC JTC1 SC37.

.
\end{IEEEbiography}

\EOD

\end{document}